\pdfoutput=1

\documentclass[11pt]{article}

\usepackage[]{ACL2023}

\usepackage{times}
\usepackage{latexsym}
\usepackage{graphicx}
\usepackage[T1]{fontenc}

\usepackage[utf8]{inputenc}

\usepackage{microtype}

\usepackage{inconsolata}

\usepackage{inconsolata}
\usepackage{graphicx}
\usepackage{booktabs}
\usepackage{multirow}
\usepackage{amsmath}
\usepackage{xcolor}
\usepackage{pdfpages}
\usepackage{afterpage}
\usepackage{stfloats}
\usepackage{xcolor}
\usepackage{amsmath}
\usepackage{xspace}
\usepackage{cleveref}
\usepackage[normalem]{ulem}
\useunder{\uline}{\ul}{}

\usepackage{amsmath}
\usepackage{algorithm}
\usepackage{algpseudocode}

\usepackage{listings} 

\usepackage{todonotes}

\newcommand{\model}{\textbf{\textcolor{darkgray}{\textsc{METAL}}}}
\newcommand{\gpt}{\textsc{GPT-4o}\xspace}
\newcommand{\llama}{\textsc{LLaMA 3.2-11b}\xspace}

\title{\model{}: A Multi-Agent Framework for Chart Generation \\ with Test-Time Scaling}

\author{
Bingxuan Li{$^{\dagger}$} \ \ \ Yiwei Wang{$^{\dagger\mathsection}$} \ \ \  Jiuxiang Gu{$^{\ddagger}$} \ \ \  Kai-Wei Chang{$^{\dagger}$} \ \ \  Nanyun Peng{$^{\dagger}$}\\
$^\dagger$  University of California, Los Angeles \quad 
$^\mathsection$ University of California, Merced  \quad 
$^\ddagger$ Adobe Research\\
\texttt{bingxuan@ucla.edu} \\\\
\href{https://metal-chart-generation.github.io}{\textcolor{magenta}{\texttt{https://metal-chart-generation.github.io}}}
}

\begin{document}

\maketitle
\begin{abstract}

Chart generation aims to generate code to produce charts satisfying the desired visual properties, e.g., texts, layout, color, and type. It has great potential to empower the automatic professional report generation in financial analysis, research presentation, education, and healthcare.
In this work, we build a vision-language model (VLM) based \textit{multi-agent} framework for effective automatic chart generation.
Generating high-quality charts requires both strong visual design skills and precise coding capabilities that embed the desired visual properties into code.
Such a complex multi-modal reasoning process is difficult for direct prompting of VLMs.
To resolve these challenges, we propose \model{} (\textbf{M}ulti-ag\textbf{E}n\textbf{T} fr\textbf{A}mework with vision \textbf{L}anguage models for chart generation), a multi-agent framework that decomposes the task of chart generation into the iterative collaboration among specialized agents. 
\model{} achieves a 5.2\% improvement in the F1 score over the current best result in the chart generation task.
Additionally, \model{} improves chart generation performance by 11.33\% over Direct Prompting with \llama.
Furthermore, the \model{} framework exhibits the phenomenon of test-time scaling: its performance increases monotonically as the logarithm of computational budget grows from $2^9$ to $2^{13}$ tokens.



\end{abstract}

\section{Introduction}
Data visualization through charts is an important part of the communication and research life cycle. Well-designed visualizations help distill complex data into digestible insights, allowing researchers, analysts, and stakeholders to identify relationships that might remain hidden in raw data \citep{qin2020making, xu2023chartbench, yang2024matplotagent}.

\begin{figure}[ht]
    \centering
    \includegraphics[width=\linewidth]{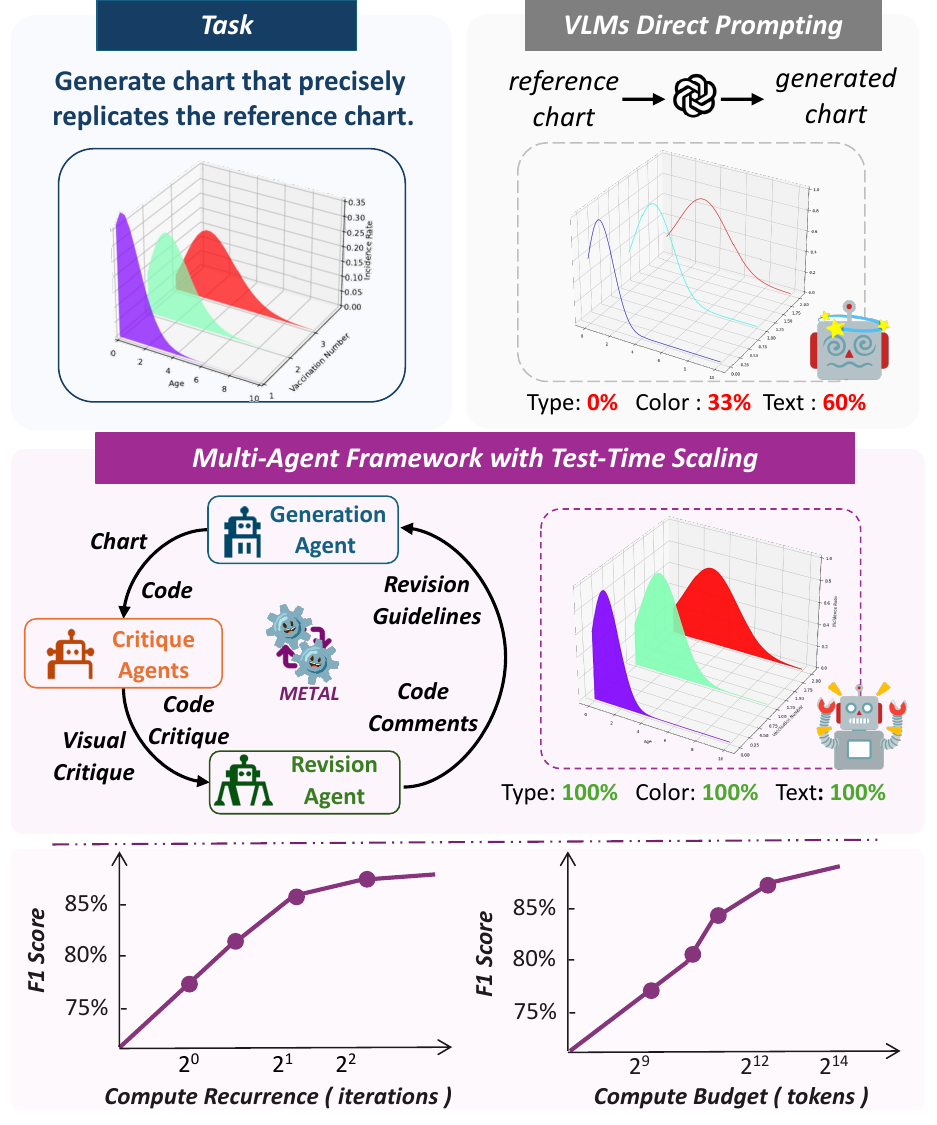}
    \caption{Direct prompting of current VLMs (e.g. \gpt ) often fails to generate charts that accurately replicate reference charts, resulting in errors in structure, color, and text alignment. Our proposed approach, \model{}, tackle this challenge with iterative refinement through generation, critique, and revision. Our experiments show that increasing the logarithm of test-time compute recurrence and token usage leads to improved performance.}
    \label{fig:teaser}
    \vspace{-0.1in}
\end{figure}

Recent advancements in vision language models (VLMs), such as GPT-4V \cite{GPT4V} and LLaVA \cite{li2024llavanext-strong}, have expanded the capabilities of language models in tackling complex multimodal problem-solving tasks. These breakthroughs have sparked growing interests in designing \textit{intelligent AI assistants} to help humans with limited coding expertise create compelling charts, leading to the emergence of a complex multi-modal generation task with crucial practical value - Chart to code generation task \citep {wu2024plot2code, han2023chartllama,  shi2024chartmimic}. 

The chart to code generation task focuses on automatically generating visualization code based on visual references. This task embodies a \textbf{highly challenging visually-grounded code generation problem} that demands robust visual understanding and advanced reasoning. The model must interpret complex visual elements—such as layouts, color schemes, and data relationships—and translate them into syntactically correct, semantically meaningful code. Successfully addressing this challenge not only improves chart replication but also paves the way for advancing the general capabilities of VLMs in multimodal learning and program synthesis.

As illustrated in Figure \ref{fig:teaser}, current state-of-the-art VLMs, such as GPT-4o, often fail to accurately interpret and reproduce the intricate visual elements and relationships embedded in reference charts.  Existing solutions, such as Best-of-N and Hint-enhanced \cite{wang2024hint}, have not effectively improved upon direct prompting of VLMs. The core challenge in leveraging VLMs for chart generation lies in effectively integrating visual comprehension with code synthesis. This complex task exceeds the capabilities of the single model or single agent.

In this paper, we present \model{}, a multi-agent framework designed for chart generation. Our framework decomposes this complex multimodal reasoning task into four specialized roles, each handled by a specialized agent: (1) \textit{Generation Agent}: Responsible for the initial translation of chart images into the corresponding code. (2) \textit{Visual Critique Agent}: Analyzes and identifies visual differences between the reference chart and the generated output. (3) \textit{Code Critique Agent}: Reviews the generated code and suggests improvements to better match the reference chart. (4)  \textit{Revision Agent}: Implements code modifications based on the combined feedback from both critique agents. During inference, these agents collaborate iteratively, critiquing and refining the code until the rendered chart achieves the desired quality. 



In contrast to existing methods, our approach delivers more concrete and targeted feedback, and iteratively refines outputs through the multi-agentic framework, leading to enhanced chart generation performance. Experiment results show that our framework achieves a 5.2\% improvement over the current best result in the chart generation task. Additionally, \model{} improves chart generation performance by 11.33\% over Direct Prompting with \llama,
demonstrating its potential to significantly enhance VLMs' ability to integrate visual understanding with code synthesis.

Furthermore, we have two key findings through in-depth analysis: (1) \textbf{Test-time scaling in the \model{} framework}: Recent works suggest the potential of scaling the number of tokens (computational budget) to enhance the reasoning performance of LLMs \citep{snell2024scaling, muennighoff2025s1}. Specifically, utilizing more tokens during inference may lead to improved performance. In this work, we found that there is a near-linear relationship between the performance and the logarithm of the computational budget in experiments. Specifically, the performance of \model{} increases monotonically as the logarithm of the computational budget grows from $2^9$ to $2^{13}$ tokens. 
(2) \textbf{Modality-tailored critiques enhance self-correction}: We observe that explicitly separating different modalities during the critique process—such as visual evaluation and code analysis—substantially enhances the multimodal self-correction capabilities of VLMs. Ablation study shows that \model{} with the separate-critique design achieves a 5.6\% improvement over the single-critique baseline.


In summary, we present \model{}, a VLMs-based multi-agent framework, which achieves significant improvements over the current best methods for chart generation, and our insights into test-time scaling and multi-modal critique offer a promising pathway for enhanced visually-grounded code generation with VLMs.

\section{Related Works}
\vspace{-0.02in}
We discuss three lines of related work: chart-to-code generation, multi-agent framework, and test-time scaling research.
\vspace{-0.01in}
\subsection{Chart Generation with VLMs}

Chart generation, or chart-to-code generation,  is an emerging task aimed at automatically translating visual representations of charts into corresponding visualization code \cite{shi2024chartmimic, wu2024plot2code}. This task is inherently challenging as it requires both visual understanding and precise code synthesis, often demanding complex reasoning over visual elements.

Recent advances in Vision-Language Models (VLMs) have expanded the capabilities of language models in tackling complex multimodal problem-solving tasks, such as visually-grounded code generation.
Leading proprietary models, such as GPT-4V~\cite{GPT4V}, Gemini~\cite{Gemini}, and Claude-3~\cite{Claude}, have demonstrated impressive capabilities in understanding complex visual patterns.
The open-source community has contributed models like LLaVA~\cite{xu2024llava-uhd, li2024llavanext-strong}, Qwen-VL~\cite{Qwen-VL}, and DeepSeek-VL~\cite{lu2024deepseekvl}, which provide researchers with greater flexibility for specific applications like chart generation.

Despite these advancements, current VLMs often struggle with accurately interpreting chart structures and faithfully reproducing visualization code. 

\begin{figure*}[ht]
    \centering
    \includegraphics[width=1\linewidth]{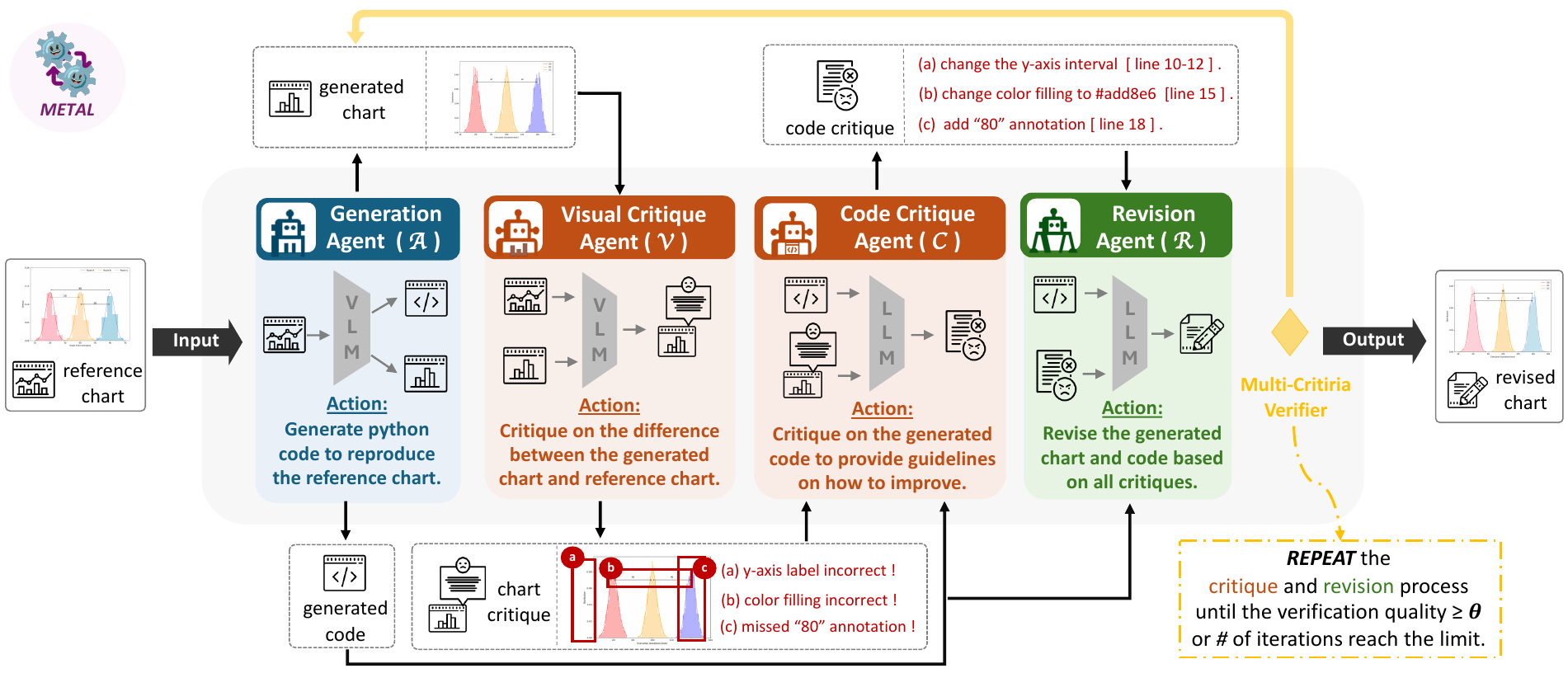}
    \caption{Overview of \model{}: A multi-agents system that consists of four specialized agents working in an iterative pipeline: (1) Generation Agent creates initial Python code to reproduce the reference chart, (2) Visual Critique Agent identifies visual discrepancies between the generated and reference charts, (3) Code Critique Agent analyzes the code and provides specific improvement guidelines, and (4) Revision Agent modifies the code based on the critiques. The process iterates until either reaching the verification score or maximum attempts limit.} 
    \vspace{-0.1in}
    \label{fig:method}
\end{figure*}

\subsection{Multi-Agents Framework}

Many researchers have suggested a paradigm shift from single monolithic models to compound systems comprising multiple specialized components~\cite{compound-ai-blog, du2024compositional}. One prominent example is the multi-agent framework.

LLMs-driven multi-agent framework  has been widely explored in various domains, including narrative generation~\cite{huot2024agents}, financial trading~\cite{xiao2024tradingagents}, and cooperative problem-solving~\cite{du2023improving}.

Our work investigates the application of multi-agent framework to the visually-grounded code generation task.

\subsection{Test-Time Scaling} 

Inference strategies have been a long-studied topic in the field of language processing. Traditional approaches include greedy decoding \cite{teller2000speech}, beam search \cite{graves2012sequence}, and Best-of-N.

Recent research has explored test-time scaling law for language model inference. For example, \citet{wu2024inference} empirically demonstrated that optimizing test-time compute allocation can significantly enhance problem-solving performance, while \citet{zhang2024scaling} and \citet{snell2024scaling} highlighted that dynamic adjustments in sample allocation can maximize efficiency under compute constraints. Although these studies collectively underscore the promise of test-time scaling for enhancing reasoning performance of LLMs, its existence in other contexts, such as different model types and application to cross-modal generation,  remains under-explored.

\section{Method}
\label{sec:method}

In this section, we introduce our method for generating precise chart representations from a given reference chart. Section~\ref{sec:task_definition} formally defines the task, Section~\ref{sec:multi_agents_system} outlines the components of our proposed approach \model{}, and Section~\ref{sec:test-time-scaling} presents the inference process of \model{}.

\subsection{Task Definition}
\label{sec:task_definition}
Given a reference chart image \(x_{\text{ref}}\) and a chart generation model, the objective is to learn the mapping
\[
    f: x_{\text{ref}} \to y,
\]
where \(y\) is a programmatic specification (e.g., Python code). When executed, \(y\) should render a chart \(O(y)\) that faithfully replicates the reference \(x_{\text{ref}}\).

\subsection{\model{}}
\label{sec:multi_agents_system}

As illustrated in Figure~\ref{fig:method}, \model{} is structured with four specialized agents  (\(G, C, V, R\)) and a multi-criteria verifier. 
All components collaborate together to iteratively refine the final output, making it more accurately replicates the reference chart. The framework is composed as follows:

\paragraph{\textbf{Generation Agent ( \(G\) )}}  
This agent is tasked to generate an initial program from the reference:
\[
    y_0 = G(x_{\text{ref}}), \quad G: \mathcal{X} \to \mathcal{Y}.
\]
This serves as the basis for further refinement.

\paragraph{\textbf{Visual Critique Agent ( \(V\) )}}  
This agent is tasked to assess the rendered chart \(O(y_t)\) against \(x_{\text{ref}}\) to detect visual discrepancies:
\[
    v_t = V(O(y_t), x_{\text{ref}}), \quad V: \mathcal{O} \times \mathcal{X} \to \mathcal{V}.
\]
Here, \(\mathcal{O}\) represents the space of visual outputs, and \(\mathcal{V}\) denotes the space of visual feedback metrics.

\paragraph{\textbf{Code Critique Agent ( \(C\) )}}  
This agent is tasked to review the generated code and provide structured critique to improve the generated code:
\[
    c_t = C(y_t), \quad C: \mathcal{Y} \to \mathcal{C}.
\]
\(\mathcal{C}\) represents the set of code critique messages ensuring correctness and efficiency.

\paragraph{\textbf{Revision Agent ( \(R\) )}}  
This agent integrates feedback from both critique agents to update the generated code:
\[
    y_{t+1} = R(y_t, v_t, c_t), \quad R: \mathcal{Y} \times \mathcal{V} \times \mathcal{C} \to \mathcal{Y}.
\]

\paragraph{\textbf{Multi-Criteria Verifier}} 
We design a heuristic-based verifier to evaluate the chart quality. The goal of verifier is to provided external source of feedback to guide four agents collaborate more efficiently.

Let \(m_j\) be the verification metrics for \(j=1,2,3\), and let \(\theta^t\) be dynamic thresholds. Then,
\[
    Q_t(O(y_t), x_{\text{ref}}) = 
    \begin{cases}
        1, & \bigwedge_{j=1}^3 m_j(O(y_t), x_{\text{ref}}) \geq \theta^t \\
        0, & \text{otherwise}.
    \end{cases}
\]

Here, we developed three verification metrics—color (\(m_1\)), text (\(m_2\)), and overall structure (\(m_3\))—to quantify the similarity between the reference and generated charts. If the generated chart meets the desired quality (i.e., the verification score of each verification metrics (\(m_1\), \(m_2\), \(m_3\)) exceeds the predefined threshold), the system triggers an early stop. More details on the implementation of the verifier, such as each verification metric, are introduced in Appendix \ref{app:verifier}.


\begin{algorithm} [tb]
\caption{Inference Procedure of \model{}}
\label{alg:inference}
\begin{algorithmic}[1]
\State \(y_0 \gets G(x_{\text{ref}})\)
\While{\(t < T_{\max}\)}
    \State \(O(y_t) \gets\) Render chart from \(y_t\)
    \State \(v_t \gets V(O(y_t), x_{\text{ref}})\) \Comment{Visual critique}
    \State \(c_t \gets C(y_t)\) \Comment{Code critique}
    \If{\(\forall j:\; m_j(O(y_t), x_{\text{ref}}) \geq \theta^t\)}
        \State \textbf{break} \Comment{Verification passed}
    \Else
        \State \(y_{t+1} \gets R(y_t, v_t, c_t)\) \Comment{Revise code}
    \EndIf
    \State \(t \gets t+1\)
\EndWhile
\State \Return \(y_t\)
\end{algorithmic}
\end{algorithm}

\subsection{Inference Procedure}
\label{sec:test-time-scaling}
Algorithm \ref{alg:inference} illustrates the inference procedure of \model{}. During inference, \model{} iteratively refines the generated code until the rendered chart meets the predefined verification threshold or reaches the maximum attempts limit $T_{\max}$. The refinement process is as follows:
\begin{align}
    y_0 &= G(x_{\text{ref}}), \\
    v_t &= V(O(y_t), x_{\text{ref}}), \\
    c_t &= C(y_t), \\
    y_{t+1} &= R(y_t, v_t, c_t).
\end{align}
The iterations terminate when
\[
    Q_t(O(y_t), x_{\text{ref}}) = 1
\]
or the max number of attempts is reached.

\section{Experiments}
\label{sec:experiments}
In this section, we systematically evaluate \model{}. Section~\ref{sec:exp_setup} details the experimental setup, Section~\ref{sec:exp_res} presents the results, and Section~\ref{sec:ablation_study} provides an ablation study to further elucidate the model's performance.

\begin{table*}[ht]
\small
\centering
\renewcommand{\arraystretch}{1.2}
\begin{tabular}{ccccccc}
\toprule
\multirow{2.3}{*}{Base Model}  & \multirow{2.3}{*}{Method} & \multicolumn{5}{c}{Average F1 Score}                                                \\ \cmidrule{3-7} 
                        &                         & \textit{Text} & \textit{Type} & \textit{Color} & \textit{Layout} & \textit{\color{purple}{Average}} \\ \midrule\midrule
\multirow{5}{*}{\llama} & Direct Prompting & 36.70\%            & 37.07\%            & 33.46\%            & 54.56\%           & \color{purple}{40.45\%}            \\ \cmidrule{3-7} 
                        & Hint-Enhanced Prompting & 38.82\%       & 38.47\%       & 36.82\%        & 51.22\%         & \color{purple}{41.33\%}          \\ \cmidrule{3-7} 
                        & Best-of-N (n = 5)        & 40.28\%       & 36.60\%       & 38.43\%        & 57.22\%         & \color{purple}{43.13\%}          \\ \cmidrule{3-7} 
                                      & \model{}  (n = 5)     & \textbf{46.69 \%↑} & \textbf{54.42\%↑}  & \textbf{47.32 \%↑} & \textbf{58.69\%↑} & \textbf{\color{purple}{51.78 \%↑}} \\ \midrule\midrule
\multirow{5}{*}{\gpt} & Direct Prompting        & 74.83\%       & 81.24\%       & 74.24\%        & 94.76\%         & \color{purple}{81.26\%}          \\ \cmidrule{3-7} 
                        & Hint-Enhanced Prompting & 77.02\%       & 80.84\%       & 72.75\%        & 93.89\%         & \color{purple}{81.12\%}          \\ \cmidrule{3-7} 
                        & Best-of-N (n = 5)        & 75.47\%       & 82.16\%       & 75.30\%        & 96.37\%         & \color{purple}{82.32\%}          \\ \cmidrule{3-7} 
                                      & \model{} (n = 5)      & \textbf{86.31\% ↑} & \textbf{84.17\% ↑} & \textbf{79.86↑}    & \textbf{95.50\%↑} & \textbf{\color{purple}{86.46\%↑}}  \\ \bottomrule
\end{tabular}
\caption{Performance comparison of \model{} and baseline methods across various base models using four evaluation metrics: Text, Type, Color, and Layout. The best performance for each metric on each base model is highlighted in \textbf{bold}. Our approach consistently outperforms the baselines, achieving the highest average F1 scores across both models, with significant improvements observed in all evaluation categories.}
\label{tbl:main_results}
\vspace{-0.1in}
\end{table*}

\subsection{Experiment Setup}
\label{sec:exp_setup}

\paragraph{Dataset} We select the ChartMIMIC dataset to evaluate \model{}. It is a benchmark that includes 1,000 human-curated (figure, instruction, code) triplets, which represent the authentic chart use cases found in scientific papers across various domains. These charts span 18 regular types and 4 advanced types, diversifying into 191 subcategories \cite{shi2024chartmimic} .

\paragraph{Automatic Evaluation Metric}
Following the approach in \cite{shi2024chartmimic}, we assess four key low-level chart elements: text, layout, type, and color. During code execution, relevant information for each element is logged for evaluation. We then compare each element in the generated chart to its counterpart in the reference chart and calculate the F1 score. Note that the evaluation metric used here \textit{differs} from the multi-criteria verification metric described in the section \ref{sec:method}.

\paragraph{Base Model}
We assess the effectiveness of \model{} on both open-source and closed-source vision-language models. Specifically, we evaluate \gpt ~\cite{gpt4o} and \llama ~\cite{llama3modelcard}. The details of model size and computation budget are introduced in Appendix \ref{app:model_size}.

\paragraph{Implementation Details}  The implementation details of each component of \model{} and the baselines are described in Appendix \ref{app:implementation}. We include the prompt templates for each VLM-driven agent and the baselines in Appendix \ref{app:prompts}.

\paragraph{Baselines}
We compare \model{} against three baseline methods, detailed as follows:
\begin{enumerate}
    \setlength{\itemsep}{0pt}
    \setlength{\parsep}{0pt}
    \setlength{\topsep}{0pt}
    \item \textit{Direct Prompting}: This baseline generates charts directly from the input prompt without any modifications or explicit guidance. It relies solely on the model's inherent ability.
    \item \textit{Hint-Enhanced Prompting}: In this approach, the input prompt is augmented with additional hints or structured guidance to help the model better understand the desired chart components \cite{wang2024hint}. Specifically, we augment the generation with a self-generated short description of a chart that provides context for elements such as layout, text, type, and color.
    \item \textit{Best-of-N}: This baseline generates multiple candidate charts in parallel, and the one that best meets a predefined verification metric is selected. We compare against Best-of-N by matching the number of iterations used in our approach.
\end{enumerate}

\subsection{Experiment Results}
\label{sec:exp_res}

The primary research question of the experiment was to assess whether our proposed method could improve the performance of the base model in the chart generation task. Table~\ref{tbl:main_results} presents the experiment result.

For the LLaMA base model, the results indicate that the performance of baseline methods varied moderately. Direct Prompting and Hint-Enhanced Prompting achieved average F1 scores of 40.45\% and 41.33\%, respectively, while Best-of-N reached 43.13\%. In contrast, \model{} yielded an average F1 score of 51.78\% with improvements observed in each metric. The average F1 score improves by 11.33\% over Direct Prompting with 5 test-time compute recurrences, which is a significant improvement.

Similarly, for the GPT base model, the baseline methods demonstrated high performance with average F1 scores ranging from 81.12\% to 82.32\%. However, \model{} outperformed these methods by a considerable margin, achieving an average F1 score of 86.46\%. Specifically, our method improved 5.2\% in average over Direct Prompting.

These results clearly demonstrate that our approach consistently improves performance across both base models and all evaluation metrics. In particular, the significant gains observed in the Text and Layout metric, along with the overall increase in average F1 scores, indicate that our model effectively captures both the structural attributes and finer details of visual data. This enhancement not only boosts the performance of both open-source and closed-source vision-language models but also maintains a high level of consistency across all metrics, underscoring the robustness and generalizability of our method.

\subsection{Ablation Study}
\label{sec:ablation_study}
To further analyze the impact of different components, we performed an ablation study by selectively removing key elements of \model{} to assess the influence of each component on the overall performance.

\paragraph{Variations Setup}
We evaluate the following variants of \model{} to assess the contribution of each agent:

\begin{itemize}
    \setlength{\itemsep}{0pt}
    \setlength{\parsep}{0pt}
    \setlength{\topsep}{0pt}
    \item \model{}$_V$: Uses only the visual critique agent, omitting the code critique component.
    \item \model{}$_C$: Uses only the code critique agent, omitting the visual critique component.
    \item \model{}$_S$: Combines the visual and code critique agents into a single, unified critique agent.
\end{itemize}

We implement each variation with \gpt as the base model. The prompt templates of each variation are attached to Appendix \ref{app:prompts}.

\paragraph{Results}

As shown in Table~\ref{tbl:ablation_study}, the full \model{} achieves the highest average F1 score, outperforming all ablated variants. Specifically, when only the visual critique agent is used (\model{}$_V$), the system obtains an average F1 score of 84.31\%, while the variant using only the code critique agent (\model{}$_C$) achieves 82.96\%. The unified agent variant (\model{}$_S$) yields the lowest average performance. 

These results indicate that both the visual and code critique agents play crucial roles in enhancing the model’s performance. The degradation observed in the ablated variants highlights that removing either component, or merging them into a single unit, compromises the system’s ability to effectively capture and refine the critical attributes of the visual data.

\begin{figure*}[ht]
    \centering
    \includegraphics[width=1\linewidth]{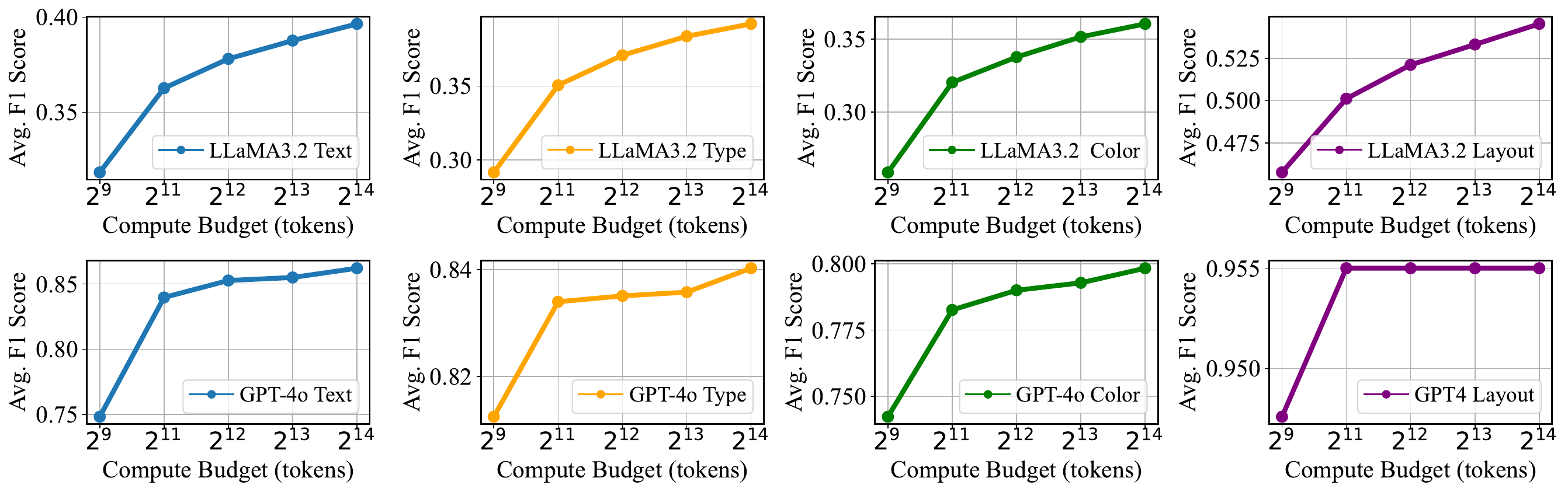}
    \caption{The performance of \model{} demonstrates an near-linear relationship with the log of compute budget.}
    \vspace{-0.1in}
    \label{fig:acc_by_iter}
\end{figure*}

\begin{table}[t]
\small
\centering
\setlength\tabcolsep{4.5pt}
\renewcommand{\arraystretch}{1.4}
\begin{tabular}{lccccc}
\toprule
\multirow{2.3}{*}{Method}      & \multicolumn{5}{c}{Average F1 Scores (\%)}                               \\ \cmidrule{2-6} 
                             & \textit{Text} & \textit{Type}& \textit{Color} & \textit{Layout} & \color{purple}{\textit{Average}} \\ \midrule\midrule
\model{}$_V$ & 83.43  & 82.57       & 77.57         & 93.69          & \color{purple}{84.31 }          \\ \midrule
\model{}$_C$ & 82.35   & 80.90             & 76.69         & 91.93          & \color{purple}{82.96}          \\ \midrule
\model{}$_S$ & 80.26  & 78.88        & 74.50         & 89.82          & \color{purple}{80.86 }          \\ \midrule\midrule
\model        & \textbf{86.31 } & \textbf{84.17 }      & \textbf{79.86 }         & \textbf{95.50 }         & \color{purple}{\textbf{86.46}}          \\ \bottomrule
\end{tabular}
\caption{Ablation study on different variants of \model{} across four evaluation metrics. The best performance for each metric is highlighted in \textbf{bold}.}
\label{tbl:ablation_study}
\vspace{-0.1in}
\end{table}

\section{Analysis and Discussion}
\label{sec:discussion}
We analyze the experimental results in this section. We highlight two interesting findings: Test-time scaling in Multi-Agent system (Section \ref{sec:result_test_time_scaling}), and modality-tailored critiques enhance the self-correction ability (Section \ref{sec:result_multimodal_self_critique}). Additionally, we discuss the advantage of \model{} (Section \ref{sec:why_metal}), and the benefit of agentic design (Section \ref{sec:result_agentic_vs_modular})

\subsection{Test-Time Scaling} 
\label{sec:result_test_time_scaling}

We investigate the relationship between the test-time computational budget and model performance. As illustrated in Figure~\ref{fig:acc_by_iter}, our analysis reveals an interesting trend:  increasing the logarithm of the computational budget leads to continuous performance improvements. This near-linear relationship indicates the test-time scaling phenomenon, demonstrating that allowing more iterations during inference could potentially enhance performance.

One potential reason for this phenomenon is the strong self-improvement capability of \model{}. Our framework is designed so that specialized agents iteratively collaborate, allowing each agent to refine its output based on feedback from others. With each iteration, errors are corrected and insights from different modalities are integrated, leading to incremental performance gains. This continual refinement process leverages the strengths of individual agents, resulting in the self-improvement capability that drives the observed performance enhancements as computational resources increase. 

Due to limited resources, we have not extended the experiment range further. However, the observed scaling implies that the framework can benefit from more iterations of collaborative self-improvement. We leave a more comprehensive exploration of this potential to the future work.


\subsection{Modality-Tailored Critiques} \label{sec:result_multimodal_self_critique}

\begin{figure} [tbp]
    \centering
    \includegraphics[width=1\linewidth]{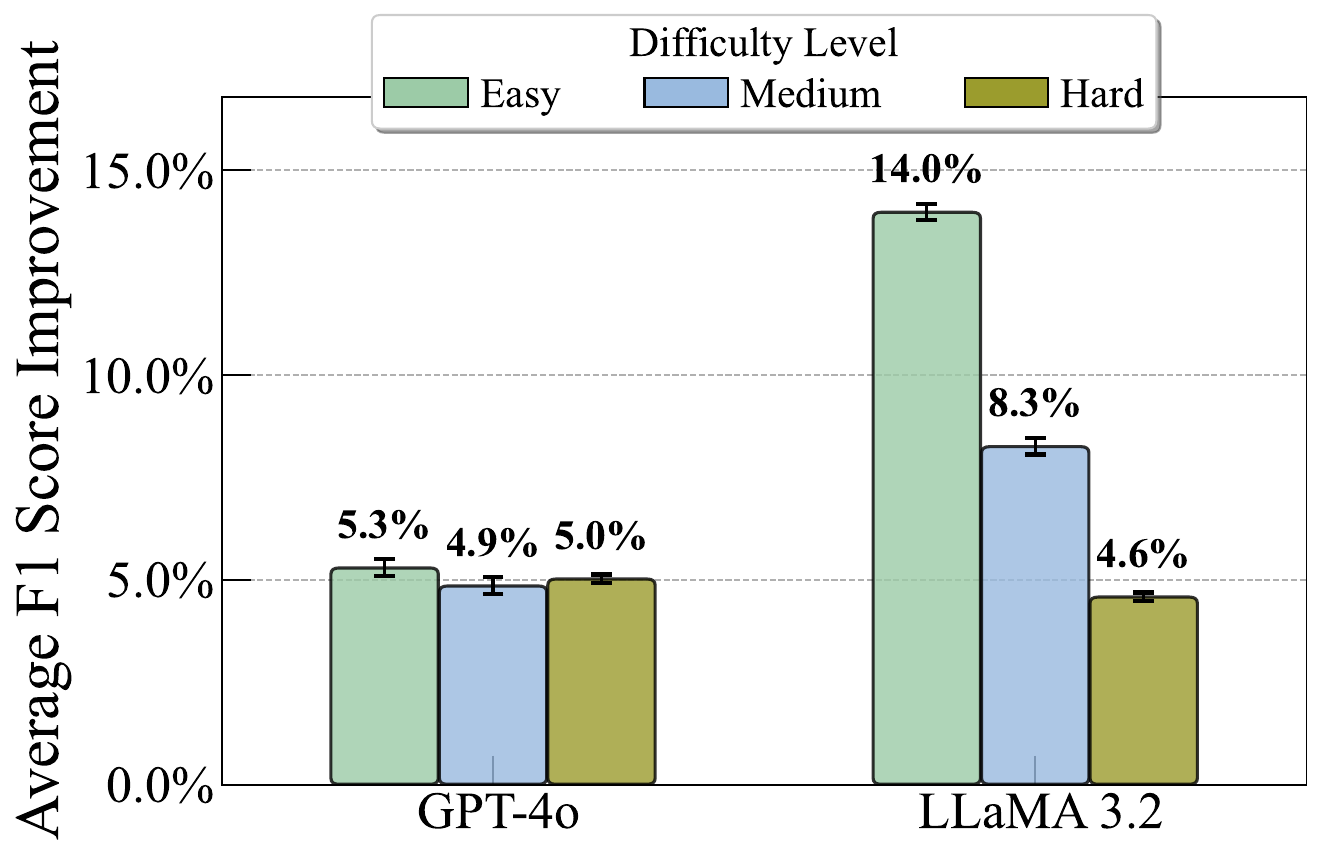}
    \caption{Performance gain after 5 compute recurrences of \model{} over different  difficulty.}
    \vspace{-0.2in}
    \label{fig:improv_by_diff}
\end{figure}

From the ablation study result shown in Table \ref{tbl:ablation_study}, we observed that separating visual and code critiques enhances the model's self-correction capabilities. In contrast, \model{}$_S$ struggles to effectively self-improve in the chart-to-code generation task.

We identify two potential reasons for this observation. First, combining both visual and code inputs results in an extended context that can overwhelm the model, leading to information loss. This dilution makes it difficult to capture key details from each modality, resulting in less accurate critiques and a reduction in overall self-correction effectiveness. Second, the self-critique process for chart generation involves distinct requirements: visual data demands spatial understanding, color analysis, and fine detail recognition, while code data requires strict adherence to syntax and logical consistency. A unified critique approach is ill-suited to address these differing needs. Without modality-specific feedback, the model struggles to detect and correct errors unique to each data type.

These findings suggest that self-correction in the multimodal context can be enhanced by leveraging tailored critique strategies for each modality.

\subsection{Why \model{}} 
\label{sec:why_metal}

We believe \model{} provides three advantages. 
First, by assigning specialized tasks to individual agents, the system effectively reduces error propagation. During inference, each agent evaluates whether to take action based on the available information and insights from other agents. This process enables each agent to serve as a safeguard, detecting and correcting mistakes before they escalate. 

Second, the modular design of \model{} enables easy modification and adaptation. For instance, one can integrate different base models tailored for specific tasks—such as employing a critique-trained model for critique agents and a generation-trained model for generation agents—to maximize overall performance. 

Third, \model{} is robust with the strong base model. Figure~\ref{fig:improv_by_diff} compares the performance of \model{} to that of Direct Prompting over five iterations across varying chart difficulty levels. \model{} with the \gpt base model achieved consistent improvements regardless of difficulty. When using \llama as the base model, the performance gains tend to diminish with increasing reference chart complexity, but the improvements remain substantial. This drop might be due to the limited critique capabilities of the \llama base model. Nonetheless, the flexibility of \model{} to replace the base model for different agents allows us to tailor the system optimally—using, for example, a critique-optimized model for critique agents and a generation-focused model for generation agents—to maximize overall performance.

\begin{figure*}[ht]
    \centering
    \includegraphics[width=1\linewidth]{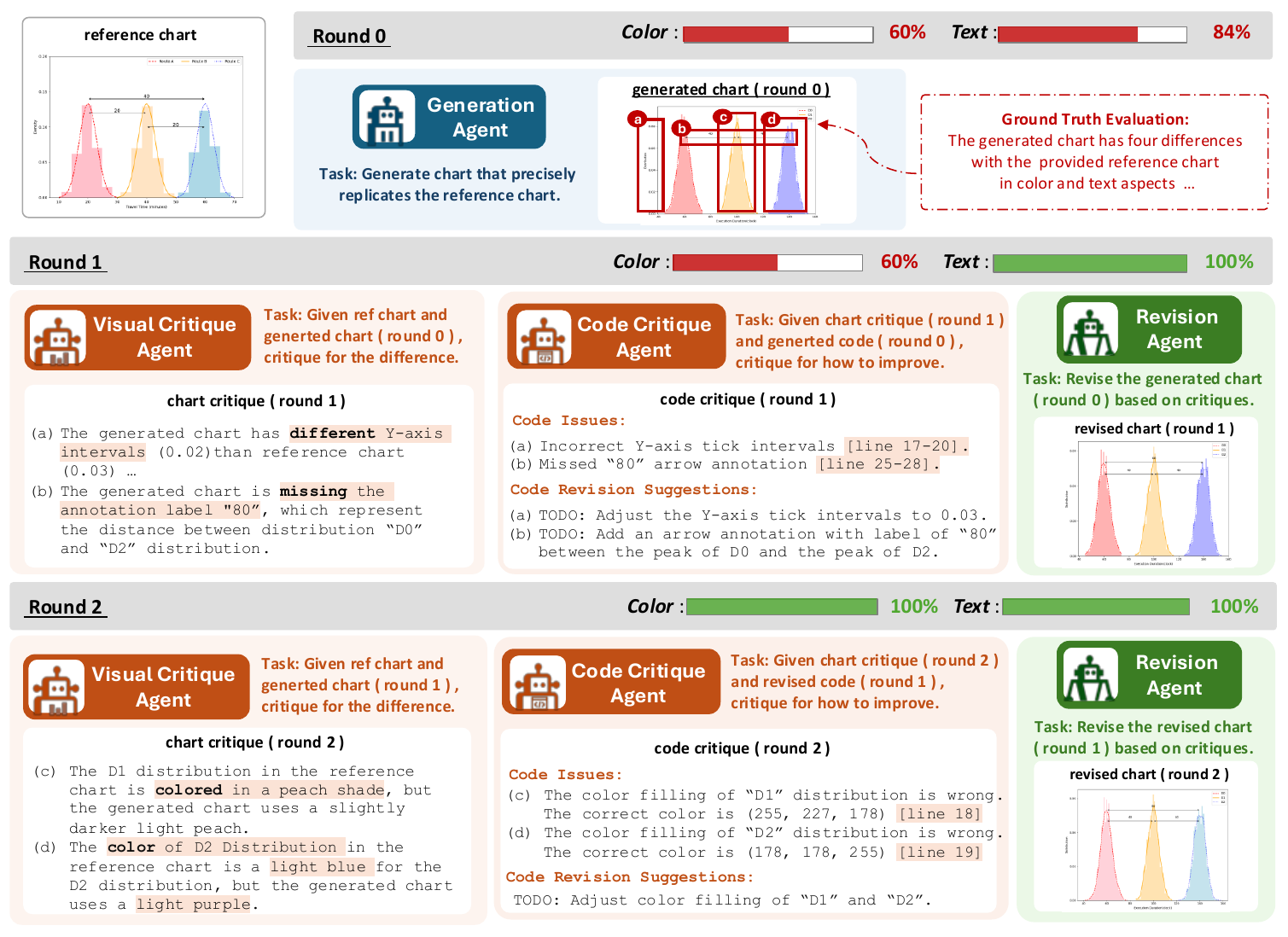}
    \caption{Case study of \model{}'s progressive refinement from initial generation to perfect. Starting from Round 0's initial generation (60\% color score , 84\% text score), the system iteratively improves the output. In Round 1, the system identifies and corrects Y-axis scale issues and missing annotations, achieving 100\% text score. Round 2 refines the color representations of distributions, achieving perfect F1 score across all metrics.}
    \label{fig:case_study}
    \vspace{-0.08in}
\end{figure*}

\subsection{Multi-Agent System vs. Modular System}
\label{sec:result_agentic_vs_modular}

We further investigate the impact of agentic behavior of \model{} on final performance. We think self-decision-making and code execution abilities are key features that distinguish the multi-agent system from a modular system. We implement a self-revision modular system without these two key abilities,  and conduct an additional ablation study on a subset of 50 data points to examine the impact of these agentic behaviors on final performance.

The results show that, compared to \model{}, there is a 4.51\% reduction in average performance gain over direct prompting. The absence of decision-making and code execution abilities in the modular system hinders its capacity to refine generated charts effectively. Specifically, the inability to execute code for chart rendering significantly diminishes the quality of the critique, and the absence of self-decision-making ability potentially leads to error propagation that further negatively impacts the self-correction process.

This comparison underscores the critical role of the agentic approach.

\section{Case Study}
\label{sec:case_study}
We perform a case study to better understand \model{}. Figure~\ref{fig:case_study} illustrates an example.  

In Round 1, two specialized critique agents analyze the generated chart. The visual critique agent detects inconsistencies in axis scaling and missing annotations, while the code critique agent identifies the corresponding code-level issues (e.g. incorrect tick intervals and absent annotations ). Based on these critiques, the revision agent modifies the chart by adjusting the Y-axis scale and adding the missing annotation. These corrections result in a significant improvement, reaching perfect text score, though color score remains unchanged.

In Round 2, the critique agents further refine the chart. The visual critique agent highlights inaccuracies in the color assignments of distributions, noting that the generated chart does not precisely match the reference chart’s colors. The code critique agent pinpoints the exact color discrepancies in the code and provides specific RGB values for correction. The revision agent incorporates these insights, adjusting the color specifications in the code. This final revision achieves perfect alignment with the reference chart, with 100\% f1 score across all evaluation metrics.

This case study demonstrates the effectiveness of \model’s multi-agent collaborative refinement process. By decomposing the task into distinct stages, \model{} can iteratively enhance the generated output. The separation of visual and code critiques ensures that both perceptual and implementation-level issues are systematically identified and addressed.

\section{Conclusion}

In conclusion, we introduce \model{}, a novel multi-agent framework that significantly enhances VLMs' performance in the chart generation task. We also reveal two interesting insights from the experiment results: the test-time scaling phenomenon in the multi-agent context, and enhanced self-correction with modality-tailored critiques.

\section*{Limitation}
Our work is not without limitations. First, our \model{} is based on VLMs, which require extensive prompt engineering. Although we selected the best-performing prompts available, it is possible that even more effective prompts could further enhance our results. Second, automatic evaluations have inherent imperfections and may not capture all details in the chart perfectly. We adopted the evaluation metric from previous work to ensure fairness. Third, \model{} has higher costs than direct prompting. Future work could explore how to optimize these costs.

\section*{Acknowledgments}
The work is partially supported by DARPA ANSR program FA8750-23-2-0004, DARPA ECOLE Program \#HR00112390060, ONR grant N00014-23-1-2780, Apple research award, and University of California at Merced.
The views and conclusions are those of the authors and should not reflect the official policy or position of DARPA or the U.S. Government.

\bibliography{custom}
\bibliographystyle{acl_natbib}

\clearpage
\section*{Appendix}
\label{sec:appendix}
\appendix
\definecolor{codegreen}{rgb}{0,0.6,0}
\definecolor{codegray}{rgb}{0.5,0.5,0.5}
\definecolor{codepurple}{rgb}{0.58,0,0.82}
\definecolor{backcolour}{rgb}{0.95,0.95,0.92}
\lstdefinestyle{mystyle}{
    backgroundcolor=\color{backcolour},
    commentstyle=\color{codegreen},
    keywordstyle=\color{magenta},
    numberstyle=\tiny\color{codegray},
    stringstyle=\color{codepurple},
    basicstyle=\ttfamily\footnotesize,
    breakatwhitespace=false,
    breaklines=true,
    captionpos=b,
    keepspaces=true,
    numbersep=5pt,
    showspaces=false,
    showstringspaces=false,
    showtabs=false,
    tabsize=2
}
\lstset{style=mystyle}

\section{Implementation}
\label{app:implementation}

In this section, we present the detailed implementation of our approach.

\subsection{VLMs driven agent Agents}
\label{app:agents}

In our implementation, we leverage VLMS to drive agents. The agents are designed to process and generate multimodal information as follows:

\paragraph{\textbf{Generation Agent and Visual Critique Agent:}}  
Both the Generation Agent and the Visual Critique Agent are designed to handle multimodal inputs. Specifically, they take as input a combination of visual data (e.g., the reference chart image or rendered chart) and textual descriptions. These agents are implemented using VLM architectures that can effectively integrate and reason over both image and text modalities. Their outputs are generated in the form of text, which provides either the initial code (in the case of the Generation Agent) or detailed visual discrepancy feedback (in the case of the Visual Critique Agent).

\paragraph{\textbf{Code Critique Agent and Revision Agent:}}  
In contrast, the Code Critique Agent and the Revision Agent are fully text-based. They accept textual inputs—either the generated code or the code accompanied by critique feedback—and produce textual outputs. Both agents are configured to generate responses up to approximately 600 tokens. 

\paragraph{\textbf{Integration of Agents:}}  
The agents interact in an iterative pipeline, where the Generation Agent first produces an initial code snippet. The Visual Critique Agent then examines the rendered output for any discrepancies relative to the reference chart, while the Code Critique Agent inspects the code for logical or syntactic issues. Finally, the Revision Agent integrates the feedback from both critique agents to modify the code. We have a Multi-Criteria Verifier (described in Appendix \ref{app:verifier} to verify the output of each iteration.

\subsection{Multi-Criteria Verifier}
\label{app:verifier}

We design three heuristic-based criteria—color, text, and overall—to assess the similarity between two images. The process begins by using EasyOCR to extract text from both the golden and generated images, and then computing a text similarity score based on the Jaccard index of the extracted text sets. In parallel, a verification from the color aspect is performed by converting the images into the HSV color space and applying predefined color ranges to count the pixels corresponding to specific colors; the resulting color histograms are compared using cosine similarity. Finally, an overall similarity measure is obtained by resizing the grayscale versions of the images and calculating the Structural Similarity Index (SSIM). The final verification result is a combination of these three metrics, providing a comprehensive assessment of image equivalence.

During the inference, the iteration will stop if the average of verification results exceeds the predefined threshold. The complete implementation code is attached as follows.
\begin{lstlisting}[language=python]
from collections import Counter
from sklearn.metrics.pairwise import cosine_similarity
from skimage.metrics import structural_similarity as ssim
import numpy as np
import cv2
import easyocr

class Verifier():
  def __init__(self, model, **kwargs):
    ocr_model_path = os.environ.get("EASYOCR_MODEL_PATH", "./easyocr_model")
    self.reader = easyocr.Reader(['en'], model_storage_directory=ocr_model_path)
  
  def extract_text(self, image_path):
        results = self.reader.readtext(image_path)
        return [text[1] for text in results]
    
  def text_similarity(self, golden_img, generated_img):
      text1 = self.extract_text(golden_img)
      text2 = self.extract_text(generated_img)
      
      intersection = len(set(text1).intersection(set(text2)))
      union = len(set(text1).union(set(text2)))
      return intersection / union if union > 0 else 0
  
  def extract_colors(self, image_path, top_n=20):
      color_ranges = {
        "red1": [(0, 100, 100), (10, 255, 255)],  # First red hue range
        "red2": [(170, 100, 100), (180, 255, 255)],  # Second red hue range
        "green": [(35, 50, 50), (85, 255, 255)],  # Green hue range
        "blue": [(100, 50, 50), (140, 255, 255)],  # Blue hue range
        "orange": [(10, 100, 100), (25, 255, 255)],  # Orange hue range
        "yellow": [(25, 100, 100), (35, 255, 255)],  # Yellow hue range
        "cyan": [(85, 100, 100), (100, 255, 255)],  # Cyan hue range
        "magenta": [(140, 100, 100), (170, 255, 255)],  # Magenta hue range
        "purple": [(125, 100, 50), (150, 255, 255)],  # Purple hue range
        "brown": [(10, 50, 20), (30, 255, 200)],  # Merged Brown & Beige range
        "pink": [(150, 100, 100), (170, 255, 255)],  # Pink hue range
        "light_blue": [(90, 50, 100), (110, 255, 255)],  # Light blue range
        "dark_blue": [(100, 150, 50), (130, 255, 150)],  # Dark blue range
        "dark_green": [(35, 100, 20), (85, 255, 120)],  # Dark green hue range
        "lime": [(40, 100, 100), (70, 255, 255)],  # Lime hue range
        "teal": [(80, 100, 100), (100, 255, 255)],  # Teal hue range
        "olive": [(30, 50, 50), (40, 255, 150)],  # Olive hue range
        "black_gray": [(0, 0, 0), (180, 50, 200)],  # Merged Black & Gray range
        "white": [(0, 0, 200), (180, 50, 255)]  # White range
     }
    
      image = cv2.imread(image_path, cv2.IMREAD_UNCHANGED)
    
      if image.shape[2] == 3:
          image = cv2.cvtColor(image, cv2.COLOR_BGR2BGRA)
      hsv_image = cv2.cvtColor(image[:, :, :3], cv2.COLOR_BGR2HSV)
      
      color_counter = Counter()
      for color_name, (lower, upper) in color_ranges.items():
          lower = np.array(lower, dtype="uint8")
          upper = np.array(upper, dtype="uint8")
          mask = cv2.inRange(hsv_image, lower, upper)
          color_counter[color_name] = cv2.countNonZero(mask)
          
      return color_counter

  
  def color_similarity(self, golden_img, generated_img):
      colors1 = self.extract_colors(golden_img)
      colors2 = self.extract_colors(generated_img)
      
      all_colors = set(colors1.keys()).union(set(colors2.keys()))
      vec1 = np.array([colors1.get(c, 0) for c in all_colors]).reshape(1, -1)
      vec2 = np.array([colors2.get(c, 0) for c in all_colors]).reshape(1, -1)
      
      return cosine_similarity(vec1, vec2)[0, 0]

  def overall_similarity(self, golden_img, generated_img):
      img1 = cv2.imread(golden_img, cv2.IMREAD_GRAYSCALE)
      img2 = cv2.imread(generated_img, cv2.IMREAD_GRAYSCALE)
      
      img1 = cv2.resize(img1, (300, 300))
      img2 = cv2.resize(img2, (300, 300))
      
      return ssim(img1, img2)
    
  def verify(self, golden_img, generated_img):
    
    text_similarity = self.text_similarity(golden_img, generated_img)
    color_similarity = self.color_similarity(golden_img, generated_img)
    overall_similarity = self.overall_similarity(golden_img, generated_img)
    
    results = {
        "text": text_similarity,
        "color": color_similarity,
        "overall": overall_similarity
    }
    
    return results
    
      
\end{lstlisting}

\section{Model Size and Computational Requirement}
\label{app:model_size}

We have developed two versions of \model{}, each built upon a different foundational model to cater to varying operational needs.

For the version using the \gpt base model, we integrate the model via the OPENAI API. In this setup, each of the four agents makes one API call per action. One single iteration—where each agent acts once—results in 4 API calls in total. In our main experiments, we perform up to 5 iterations per trial.

Alternatively, the \llama-based version of \model{} is hosted locally on two NVIDIA A100 Tensor Core GPUs, each with 40 GB of GPU memory. Each of the four agents runs its own instance of the \llama model, leading to an overall GPU memory requirement of approximately 70 GB.

\section{Prompt Templates}
\label{app:prompts}

\subsection{\model{}}

This section lists all prompt templates used in \model{}.

\begin{lstlisting}[language=python]
## Generation Agent ##
# Note: The generation prompt template is adapted from ChartMIMIC.
generation_prompt_template = "You are an expert Python developer who specializes in writing matplotlib code based on a given picture. I found a very nice picture in a STEM paper, but there is no corresponding source code available. I need your help to generate the Python code that can reproduce the picture based on the picture I provide. \nNote that it is necessary to use figsize={figsize} to set the image size to match the original size. \nNow, please give me the matplotlib code that reproduces the picture below."

## Visual Critique Agent ##
visual_critique_prompt_template = "You are a professional data scientist tasked with evaluating a generated chart against a reference chart.

Objective: 
Please identify whether there is (are) issue(s) in the generated chart that diverge from the reference chart concerning the {lowest_metric}, and give concrete feedback.
Compare the original chart (left) and the generated chart (right) in the provided image.
Provide a detailed critique of how the generated chart diverges from the reference chart concerning the {lowest_metric}.
Avoid commenting on unrelated metrics or general stylistic choices unless they directly affect the {lowest_metric}.
Only critique on elements that in reference chart but not in generated chart. 
For example, if the reference chart doesn't have a title, don't critique the title in the generated chart.

Instructions:
{instructions}

Response Format:
1. Observation (Reference Chart): Identify the chart elements in the reference chart.
2. Observation (Generated Chart): Identify the chart elements in the generated chart.
3. Critique: Issues in the generated chart that diverge from the reference chart concerning the {lowest_metric}. Be specific and detailed. If there is numeric value or text, please provide the exact value or text.

Note: If you believe there is no issue, please respond with SKIP.
"

## Code Critique Agent ##
code_critique_prompt_template = "You are a professional data scientist tasked with analyzing input code and adding targeted TODO comments based on the provided critique.

Instructions:
Please identify whether there is(are) issue(s) in the input code that need to be addressed based on the visual critique, and give concrete feedback.
Identify the specific issues raised in the critique.
Offer clear and actionable suggestions to address the identified issues.
Insert TODO comments directly into the input code to indicate necessary changes.
Place the TODO comment above the line of code that requires modification.
Be specific and practical in the TODO comments. Avoid generic suggestions or additions unrelated to the critique.
Do not make changes beyond adding TODO comments to the code, such as change existing code or add new lines of code.
Do not modify the code or add comments about code style, unrelated improvements, or hypothetical enhancements outside the critique's scope.
Do not mention reference charts in the TODO comments, making the comment self-contained.

Response Format:
1. Issues: Summarize the issues identified in the critique.
2. Suggestions: Provide specific suggestions for addressing the issues.
3. Full Code with added TODO Comments: Present the input code with the TODO comments added above the relevant lines. Please ONLY add TODO comments to the input code, do not modify the code in any other way.

Critique:
{critique}

Code to Comment On:
```python
{code}
```

Note: If you believe there is no issue, please respond with SKIP.
"

## Revision Agent ##
code_revision_prompt_template = "You are a professional data scientist tasked with revising the input code.

Objective:
The input code contains TODO comments that need to be addressed.
Please carefully review the code and make the necessary revisions to address the TODO comments.
Each comment might need more than one line of code to address.
Match the other lines of code style and structure in the input code.
Ensure that the revised code is correct and functional.
Return the FULL revised code to ensure the code is ready for the next stage of development. 

Response Format:
Full Code of the Revised Version: Present the revised code with the changes made to address the TODO comments.

Code to Revise:
```python
{code}
```
"
\end{lstlisting}

\subsection{Variations}

This section lists all prompt templates used in variations from the ablation study.
\begin{lstlisting}[language=python]
# Variations
# 1. Metal-s: GenerationAgent, SingleCritiqueAgent, RevisionAgent, VerificationAgent
# 2. Metal-v: GenerationAgent, VisualCritiqueAgent, VisualRevisionAgent, VerificationAgent
# 3. Metal-c: GenerationAgent, TextCritiqueVisualAgent, RevisionAgent, VerificationAgent


## SingleCritiqueAgent (Metal-s) ##
SingleCritiqueAgent_prompts_template = "You are a professional data scientist tasked to critique the generated chart against a reference chart to improve the code for generating the chart.

Objective: 
There is (are) issue(s) in the generated chart that diverge from the reference chart regarding the {lowest_metric}. Compare the original chart (left) and the generated chart (right) in the provided image.
Observe the differences between the reference chart and the generated chart, and provide a detailed critique of the generated chart.
Insert TODO comments directly into the input code to indicate necessary changes. Place the TODO comment above the line of code that requires modification. Be specific and practical in the TODO comments. Avoid generic suggestions or additions unrelated to the critique.

Instructions:
- Step1-2: 
Avoid commenting on unrelated metrics or general stylistic choices unless they directly affect the {lowest_metric}. 
Only critique on elements that in reference chart but not in generated chart. 
For example, if the reference chart doesn't have a title, don't critique the title in the generated chart.
Here are insturction for the chart critique:
{chart_instructions}
- Step3-4:
If the chart critique specifies particular color values, include a TODO comment in the code to remove the opacity setting (e.g. alpha). This ensures the color is accurately replaced with the specified values from the critique
If the chart critique identifies incorrect or missing text (of label, annotation, etc.), please include the correct text in the TODO comment for easy reference.
If the chart critique identifies the mismatched type of chart, please add TODO comments above the type-related functions calls to correct the chart type. 

Response Format:
1. Chart Critique: Issues in the generated chart that diverge from the reference chart. Be specific and detailed.
2. Code Critique: Provide a critique of the code that generated the chart. Identify the issues and suggest improvements by adding TODO comments.
3. Full Code with added TODO Comments: Present the input code with the TODO comments added above the relevant lines. Do not modify the code directly.

Code to Comment On:
```python
{code}
```
"

## VisualRevisionAgent (Metal-v) ##
VisualRevisionAgent_prompt_tempate ="
You are a professional data scientist tasked with revising the input code based on the visual critique provided.

Objective:
The input code contains a Python script that generates a chart. 
The chart is intended to replicate the reference image, but there could be discrepancies between the two.
The chart that generated by input code has been reviewed by a data visualization expert who provided a visual critique. 
Your task is to revise the code to address the issues identified in the critique and improve the chart accordingly.
Identify the discrepancies and issues in the generated chart based on the critique.
Make necessary modifications to the input code to resolve the identified issues and improve the chart.
Match the existing style and structure of the input code. 
Ensure correctness and functionality.
Return the FULL revised code to ensure it is ready for the next stage of development.

Inputs:
- Visual Critique:
{visual_critique}

- Code to Revise:
```python
{code}
```
"""

## TextCritiqueVisualAgent (Metal-c) ##
TextCritiqueVisualAgent_prompt_template = "You are a professional data scientist tasked with analyzing input code and adding targeted TODO comments based on the reference image.

Instructions:
The input code contain python script to generate a chart.
The chart generated by the input code should replicate the reference image provided.

There is(are) issue(s) in the input code that need to be addressed to match the reference image.
Identify the specific issues raised in the input code that prevent it from replicating the reference image.
Offer clear and actionable suggestions to address the identified issues.
Insert TODO comments directly into the input code to indicate necessary changes.
Place the TODO comment above the line of code that requires modification.
Be specific and practical in the TODO comments. Avoid generic suggestions or additions unrelated to the critique.
Do not make changes beyond adding TODO comments to the code, such as change existing code or add new lines of code.
Do not modify the code or add comments about code style, unrelated improvements, or hypothetical enhancements outside the critique's scope.
Do not mention reference charts in the TODO comments, making the comment self-contained.

Response Format:
1. Issues: Summarize the issues identified.
2. Suggestions: Provide specific suggestions for addressing the issues.
3. Full Code with added TODO Comments: Present the input code with the TODO comments added above the relevant lines. Please ONLY add TODO comments to the input code, do not modify the code in any other way.

Code to Comment On:
```python
{code}
```
"
\end{lstlisting}

\subsection{Baselines}

This section lists all prompt templates used in baselines.
\begin{lstlisting}[language=python]
## HintEnhanced Baseline ##
# Note: This prompt template is adapted from ChartMIMIC.

hint_enhanced_prompt_template = "You are an expert Python developer who specializes in writing matplotlib code based on a given picture. I found a very nice picture in a STEM paper, but there is no corresponding source code available. I need your help to generate the Python code that can reproduce the picture based on the picture I provide.\n\nTo ensure accuracy and detail in your recreation, begin with a comprehensive analysis of the figure to develop an elaborate caption.\nThis caption should cover, but not be limited to, the following aspects:\n1. Layout Analysis: e.g., identify the picture's composition, noting the presence and arrangement of any subplots.\n2. Chart Type Identification: e.g., determine how many charts within a subplot. Are they independent, or do they share a common axis?\n3. Data Analysis: e.g., summarize the data trend or pattern.\n4. Additional Features: e.g.,identify any supplementary elements such as legends, colormaps, tick labels, or text annotations that contribute to the figure's clarity or aesthetic appeal.\n\nNow, given the picture below, please first output your comprehensive caption and then use the caption to assist yourself to generate matplotlib code that reproduces the picture.\nNote that it is necessary to use figsize=({figsize}) to set the image size to match the original size."


\end{lstlisting}

\end{document}